\pdfoutput=1

\documentclass[11pt]{article}

\usepackage[]{emnlp2021}

\usepackage{times}
\usepackage{latexsym}

\usepackage[T1]{fontenc}

\usepackage[utf8]{inputenc}

\usepackage{times}
\usepackage{latexsym}
\usepackage{soul}
\usepackage{tabulary,multirow,overpic,xcolor}
\usepackage{booktabs}
\usepackage{balance}

\newcolumntype{x}[1]{>{\centering\arraybackslash}p{#1pt}}

\newlength\savewidth\newcommand\shline{\noalign{\global\savewidth\arrayrulewidth
		\global\arrayrulewidth 1pt}\hline\noalign{\global\arrayrulewidth\savewidth}}

\newcommand{\eg}{\textit{e}.\textit{g}.}
\usepackage{microtype}

\newcommand{\app}{\raise.17ex\hbox{$\scriptstyle\sim$}}

\usepackage{amsfonts}
\usepackage{url}
\usepackage{xcolor}
\usepackage[ruled,linesnumbered]{algorithm2e}
\usepackage{graphicx}
\usepackage{caption}
\usepackage{subcaption}
\usepackage{bbm}

\usepackage{amsmath,bm}
\DeclareMathOperator*{\argmax}{arg\,max}

\usepackage{microtype}

\title{VideoCLIP: Contrastive Pre-training for\\Zero-shot Video-Text Understanding}

\date{}

\author{Hu Xu$^{1}$, Gargi Ghosh$^{1}$, Po-Yao Huang$^{12}$, Dmytro Okhonko$^{1}$, Armen Aghajanyan$^{1}$\\ 
\textbf{Florian Metze,}$^{1}$ \textbf{Luke Zettlemoyer}$^{1}$ \and \textbf{Christoph Feichtenhofer}$^{1}$\\
$^1$Facebook AI\\
$^2$Carnegie Mellon University\\
\texttt{\{huxu,gghosh,berniehuang,oxo,armenag}\\
\texttt{fmetze,lsz,feichtenhofer\}@fb.com}
}

\begin{document}
\maketitle
\begin{abstract}
\vspace{-10pt}

We present VideoCLIP, a contrastive approach to pre-train a unified model for zero-shot video and text understanding, without using any labels on downstream tasks. \mbox{VideoCLIP} trains a transformer for video and text by contrasting temporally overlapping positive video-text pairs with hard negatives from nearest neighbor retrieval. Our experiments on a diverse series of  downstream  tasks, including sequence-level text-video retrieval, VideoQA, token-level action localization, and action segmentation reveal state-of-the-art performance, surpassing prior work, and in some cases even outperforming supervised approaches. Code is made available at \url{https://github.com/pytorch/fairseq/tree/main/examples/MMPT}. 
\end{abstract}

\section{Introduction}
\label{sec:intro}

The popular ``pre-training + fine-tuning'' paradigm has revolutionized NLP ~\cite{devlin-etal-2019-bert,liu2019roberta,yang2019xlnet,lewis-etal-2020-bart} and CV ~\cite{chen2020simple,he2020momentum} over the last few years.
Although models trained this way can achieve impressive performance, they still require task-specific annotated data and fine-tuning for each end task. 
Recent work adopt pre-training for zero-shot transfer to end tasks without fine-tuning, including GPT~\cite{radford2018improving,radford2019language,brown2020language} for NLP tasks and CLIP~\cite{radford2021learning} for image classification.


\begin{figure}[!t]
\centering
\vspace{-10pt}
\includegraphics[width=\linewidth]{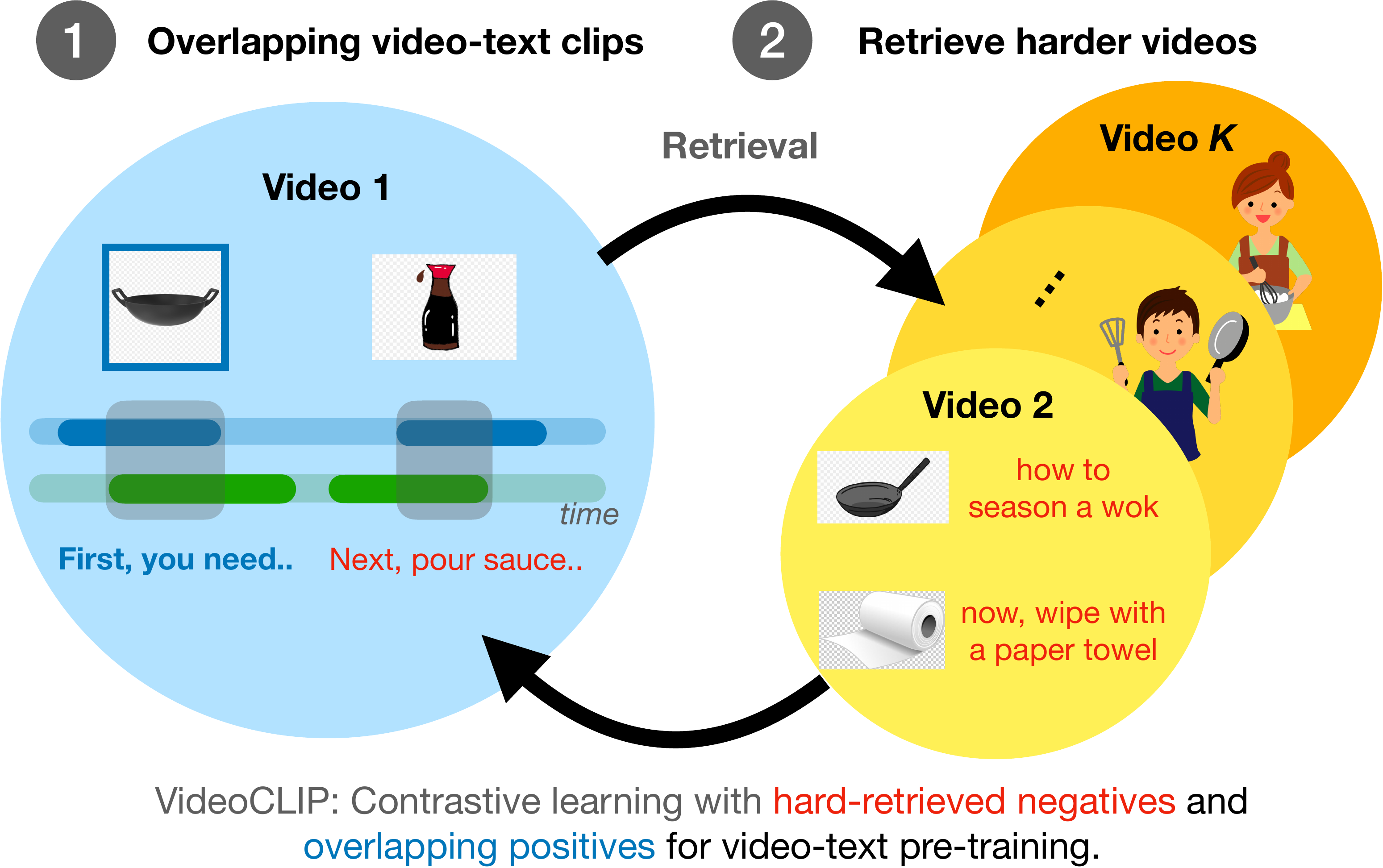}
\vspace{-15pt}

    \caption{VideoCLIP aims for zero-shot video understanding via learning fine-grained association between video and text in a transformer using a contrastive objective with two key novelties: (1) for \textit{positive} pairs, we use video and text clips that are \textit{loosely} temporarily overlapping instead of enforcing strict start/end timestamp overlap; (2) for \textit{negative} pairs
    , we employ a retrieval based sampling technique that uses video clusters to form batches with mutually harder videos.} \vspace{-15pt}
\label{fig:overview}
\end{figure}

This paper focuses on pre-training for zero-shot transfer to \textit{video}-text understanding tasks.
Our approach pre-trains a Transformer model~\cite{vaswani2017attention,devlin-etal-2019-bert} with a contrastive objective~\cite{oord2018representation,chen2020simple} using pairs of video-text clips. 
Different from CLIP that scales pre-training data for zero-shot transfer to image classification on an explicitly assembled dataset using a simple contrastive objective~\cite{chen2020simple}, this paper uses a \textit{publicly established} pre-training dataset, HowTo100M~\cite{miech2019howto100m}, for~\textit{zero-shot video understanding}. 
We show that the resulting pre-trained model can be either directly applied to, or fine-tuned on, a series of video-text tasks at both the global sequence and local clip/token level.
 

We find that straightforward objectives~\cite{chen2020simple} lead to poor results, and hypothesize that learning fine-grained associations between video and text is crucial for success of zero-shot transfer to end tasks. Since end tasks may require different granularities of video-text correspondence. The granularity can be about sequence length (such as long video versus short text (\eg classification), token level or sequence level) and semantics (``apple'' vs ``banana'' or ``apple'' vs ``car''). Previous efforts sample short, temporally aligned video and text clips with contrastive learning within a random batch, falling short on learning the fine-grained association between video frames and word tokens.

We present VideoCLIP 
that 
aims to pre-train a \textit{unified} video-text representation with contrastive learning using two key techniques (see~Fig.~\ref{fig:overview}) to compute the training objective. 

First, 
we aim to improve the association of video and text with different sequence lengths.
Although the majority of video clips and text transcriptions are not semantically aligned~\cite{miech2019howto100m}, current video-text models are trained with exact temporal alignment.
As a result, multiple or longer text clips may have better alignment with a video clip~\cite{miech2020end} and many
clips may not have any corresponding captions (see a detailed discussion of issues in \S \ref{sec:overlap}).
To address these issues, we pre-train with temporally \textbf{overlapped} pairs of video and text clips (of varying length), thereby greatly increasing the quality and quantity of the video-text alignment. 
We show in experiments that this simple and general approach significantly improves performance.

Second, we learn fine-grained video-text similarity from a contrastive loss with a new method for gathering (implicitly) harder negative pairs.
Although existing works contrast intra-video clips via sampling multiple clips from the same video \cite{miech2019howto100m,miech2020end}, we find that mining clips from other videos can provide much more challenging negatives.
We propose a \textbf{retrieval augmented pre-training} approach to retrieve a cluster of videos that are similar to each other for each training batch.
Retrieval-augmented pre-training alternatively performs retrieving video clusters and uses the retrieved video clusters for pre-training
(see \S~\ref{sec:retri} for details).

After pre-training, we apply our model for zero-shot transfer \textit{without} any fine-tuning on target dataset labels. We directly use our pre-trained model on a diverse set of~\textit{four} tasks in ~\textit{five} datasets, including text-video retrieval (for text-to-video similarity), VideoQA (for video-to-text similarity), 
action localization (for video frame to text label similarity) and segmentation (for video token to text label similarity with rejection) (see \S\ref{sec:zs}).

Our experiments reveal that VideoCLIP has strong performance, even compared to supervised approaches which use human-annotated labels on the downstream tasks. For example, in text-video retrieval on Youcook2~\cite{zhou2017towards}, VideoCLIP outperforms all existing zero-shot methods and even outperforms fully supervised pre-training + fine-tuning methods, but without using any labels.

In summary, the main contributions of this paper include: 
\textit{(i}) we propose to pre-train a \textit{unified} model that is capable of zero-shot transfer to \textit{multiple} end tasks for video-text understanding, even surpassing fully-supervised methods in some cases, 
and (\textit{ii}) we introduce two novel techniques to improve the learning of fine-grained video-text association.

\vspace{5pt}
\section{Related Work}
\label{sec:related_work}
\paragraph{Pre-training for Zero-shot Transfer.}
Recently, the paradigm of pre-training has made impressive progress with the scale of training data and computational power. 
For example, in NLP, the paradigm has shifted from learning word embeddings for task-specific architecture \cite{mikolov2013distributed,bojanowski2017enriching,peters2018deep}, to pre-training+fine-tuning~\cite{devlin-etal-2019-bert,liu2019roberta,lewis-etal-2020-bart} and few-shot/zero-shot transfer~\cite{radford2018improving,radford2019language,brown2020language,alayrac2020self,ramesh2021zero}
that have task-agnostic architecture.
One line of pre-training for zero-shot transfer focuses on generative (auto-regressive) models~\cite{radford2018improving,radford2019language,brown2020language}, where examples and prompts of an end task are used as context for a language model to respond properly to that task~\cite{brown2020language}; the other line of studies focuses on discriminative models~\cite{alayrac2020self,miech2020end}, where a   similarity search or ranking model learns a joint space (\eg ~via contrastive learning~\cite{chen2020simple,he2020momentum}) and later transfer to a particular task.
Recently, CLIP~\cite{radford2021learning} transfers image-text similarity to many image classification tasks, where the text branch serves as supervision for learning a general image representation and subsequently serves as a hyper network for downstream vision tasks.
Our effort aligns with the latter line of work, but is the first to  transfer a pre-trained discriminative model to a broad range of tasks in multi-modal video understanding.


\paragraph{Multi-modal Video-Text Pre-training.}
Multi-modal models have also adopted the pre-training+fine-tuning paradigm.
One line of work 
adopts multiple unimodal encoders for retrieval tasks.
For example, \cite{miech2019howto100m,miech2020end,ging2020coot,gabeur2020multi,alayrac2020self,patrick2021supportset,huang21naacl} adopt contrastive learning for pre-training and shows the possibility of zero-shot transfer to text-video retrieval tasks.
CBT \cite{sun2019contrastive}, HERO \cite{li-etal-2020-hero}, VideoAsMT \cite{korbar2020video} and UniVL \cite{luo2020univilm} adopt multi-task learning (MTL) for pre-training on retrieval tasks.
HERO \cite{li-etal-2020-hero} and UniVL \cite{luo2020univilm} further adopt a cross-encoder to further learn the fusion of different modalities.

The other line of work adopts a single \textit{cross-modal} encoder and concatenates the vision and text sequences as inputs, including
VideoBERT \cite{sun2019videobert}, Unicoder-VL \cite{li2020unicoder}, VL-BERT \cite{Su2020VL-BERT:}, UNITER \cite{chen2020uniter}, VLP \cite{zhou2018end},
ActBERT \cite{zhu2020actbert} and VLM \cite{xu-etal-2021-vlm}.
Although this approach is intuitive, it limits the capability of zero-shot transfer. For example, it is non-trivial to perform retrieval tasks on a single encoder as feeding vision and text in a pairwise manner is not flexible and data efficient~\cite{luo2020univilm}.
\noindent\textbf{Retrieval Augmented Training.}
Augmenting traditional training with a non-parametric retrieval component has recently shown impressive results in pre-training
\cite{khandelwal2019generalization,guu2020realm,lewis2020pre}
and QA
\cite{izacard2020leveraging,karpukhin2020dense}. 
We find that contrastive learning and retrieval augmented training can have good synergy because the former aims to discriminate examples and the latter aims to find harder examples for discrimination. 
To the best of our knowledge, there is no existing work of retrieval augmented training for video, perhaps because videos exhibit unique challenges for data-efficient training (see \S\ref{sec:retri}).

\section{VideoCLIP Pre-training}
\label{sec:pretrain}

In the paradigm of multi-modal video-text pre-training for zero-shot transfer, the key challenge is to learn fine-grained association in-between video and text to cover the diverse needs of end tasks.
We cover VideoCLIP pre-training in this section, and discuss the needs of zero-shot transfer to different end tasks in the next section.
We first describe video and text model backbone and contrastive loss;
then we propose overlapped video and text clips to improve the association of positive pairs; lastly, we describe retrieval augmented pre-training to improve the mining of negative examples.

\subsection{Video and Text Encoding}
VideoCLIP consumes pairs of video and text clips $(v, t)$ as inputs.
It makes no assumptions on the encoder architectures and can work with any video and text backbone.
We use Transformer~\cite{vaswani2017attention} model for both the video and text.
The video features, extracted by a convolutional neural network (CNN), are first projected to \textit{video tokens} before fed into our video transformer, as described next.

\paragraph{Video and Text Transformers.}
Let $\bm{c}_v$ be a video clip of a sequence of continuous frames (we use  \textit{bold} symbols to indicate sequences). We feed $\bm{c}_v$ into a (frozen) pre-trained video encoder $f_{\theta_\text{CNN}}$ and then apply a trainable MLP, $f_{\theta_\text{MLP}}$, with weights $\theta_\text{MLP}$ to obtain \textit{video token}s  $\bm{x}_v \in \mathbb{R}^d$ with the same embedding dimension, $d$, as for word embeddings in our architecture:
\begin{equation}
\begin{split}
\bm{x}_v = f_{\theta_\text{MLP}}(\texttt{stopgrad}(f_{\theta_\text{CNN}}(\bm{c}_v))), 
\end{split}
\end{equation}
where $\texttt{stopgrad}$ is a stop-gradient operation, to reflect that the video CNN is frozen. 

Similarly, vectors for text tokens $\bm{x}_t$ are obtained via embedding lookup as in BERT~\cite{devlin-etal-2019-bert}.
Then 
$\bm{x}_v$ and $\bm{x}_t$ are feed into 
two separate trainable Transformers,
$f_{\theta_\text{v}}$ and $f_{\theta_\text{t}}$, respectively,
to obtain the hidden states for video and text tokens
\begin{equation}
\label{eq:forward}
\begin{split}
\bm{h}_v =  f_{\theta_\text{v}}(\bm{x}_v), \bm{h}_t =  f_{\theta_\text{t}}(\bm{x}_t).
\end{split}
\end{equation}
To obtain the hidden states (\textit{i.e.}~global features) of video and text clips, we apply average pooling over the sequence of tokens for video and text, respectively
\begin{equation}
\begin{split}
{z}_v = \texttt{AvgPool}(\bm{h}_v), {z}_t = \texttt{AvgPool}(\bm{h}_t).
\end{split}
\end{equation}
We use average pooling (instead of using the \texttt{[CLS]} token) to encourage $f_{\theta_\text{v}}$ and $f_{\theta_\text{t}}$ to learn token-level representations that may benefit token-level tasks, such as action localization and action segmentation (see Section~\ref{sec:zs}).

VideoCLIP aims at pre-training the \textit{unified} video-text representation, captured by the 
Transformer model parameters $\theta_\text{v}$ and $\theta_\text{t}$ for video and text, and consequently use it for zero-shot downstream tasks. 
In appendix, we also explore \textit{shared weights} for video and text,  $\theta_\text{v} \equiv \theta_\text{t}$, and our ablations show that separate video/text transformers yields slightly better performance.

Notably, using a frozen video backbone ($f_{\theta_\text{CNN}}$) enables us to go beyond short-term visual input (typical video CNNs~\cite{xie2018rethinking,feichtenhofer2019slowfast} only capture temporal windows of \app3 seconds), and allows us to model \textit{long-term visual-textual correspondences} spanning \app32 seconds. We describe our training methodology next. 

\subsection{Contrastive Loss}
We use a contrastive loss (\mbox{InfoNCE} \cite{oord2018representation} objective) to learn the correspondence between video and text. 

In particular, we minimize the sum of two multi-modal contrastive losses:
\begin{equation}
\small
\label{eq}
\mathcal{L}=-\sum_{(v, t)\in B}\Big(\log\text{NCE}(z_v, z_t)
+ \log \text{NCE}(z_t, z_v)\Big),
\end{equation}
where $B$ is 
the batch 
that contains sampled video-text pairs and $\text{NCE}(z_v, z_t)$ and $\text{NCE}(z_t, z_v)$ corresponds to the contrastive loss on video-to-text similarity and text-to-video similarity. Specifically, the video-to-text contrastive loss is given by

\begin{equation}\label{eq:nce} 
 \text{NCE}(z_v, z_t) = {\frac{ \exp\left({z_v \cdot z_t^+ / \tau}\right)}{ {\sum_{z \in \{z_t^+, z_t^-\}}} {\exp\left({z_v \cdot z / \tau}\right)}   }},
\end{equation}
with $\tau$ being a temperature hyper-parameter and $z_{t}^{+}$ are \textit{positive} embedded text clips overlapping with video clip embedding $z_v$, and  $\{z_{t}^{-}\}$ are \textit{negative} embedded text clips that are implicitly formed by other text clips in the training batch.
The text-to-video loss $\text{NCE}(z_t, z_v)$ is defined symmetrically. 
The next sections (\S\ref{sec:overlap} and \S\ref{sec:retri}) describe how we construct the positive, $z_t^{+}$,  and  negatives,  $\{z_t^{-}\}$, in our pre-training objective \eqref{eq:nce}.

\subsection{Overlapped Video-Text Clips}
\label{sec:overlap}

To build overlapping \textit{positive} video/text pairs, we 

(\textit{i}) sample a text clip (because sampling a video clip first may not have nearby corresponding text); 

(\textit{ii}) sample a timestamp within the boundary of text clip as the \textit{center} for a video clip; 

(\textit{iii}) grow a video clip with random duration (up to \app32 seconds) from this center timestamp.

Our empirical results show this simple method works well in practice, and we discuss its benefits w.r.t.~prior efforts next.

\noindent \textbf{Low Relevance Temporal Alignment.}
Existing video-text pre-training methods, \eg,~\cite{miech2019howto100m}, consider temporally exactly aligned clips (video and text clips sharing the same start/end timestamps).
Although strict alignment seems natural, it is less likely that temporally aligned video and text clips are also semantically close in short clips.
For example, a video clip of ``\textit{a person speaking}'' may have a low relevance\footnote{We use the term low relevance instead of noisy alignment because temporally aligned clips may still have low relevance on certain perspectives, such as positive emotions, an opened mouth with any transcription popping up, and ``going to'' in transcription indicates visual contents may show up later.}
with the exact temporally aligned transcription ``\underline{I am going to show you how to cook fried rice}''. 
However, a later video clip showing ``\textit{rice in wok}'' may have a better semantic visual alignment.
One explanation for this low relevance of temporal alignment is that humans are less likely to speak and perform actions simultaneously.

Using exact temporal alignment limits the examples considered in the contrastive loss.
Taking the previous $\text{NCE}(z_v, z_t)$ term as an example, the low relevance (positive) pair could be in the numerator of the objective \eqref{eq:nce}, whereas higher relevance pairs (\eg~\textit{rice in wok} appearing later in a video with an introductionary text clip of ``{I am going to show you how to cook fried rice}'') are possibly used as negative pairs, under exact temporal alignment for constructing positive/negative samples.
Although existing work \cite{miech2020end} aligns multiple nearby text clips with one (short) video clip of fixed 3.2 seconds duration, this only partially solves the low relevance problem and can attenuate noise, as the text clips may only partially correspond to the visuals and might have no temporal overlap with the short-duration video clip per se.

\noindent \textbf{Better Video-Text Association.}
As such, we believe a (self-supervised) method that can curate higher relevance video-text pairs at a large-scale is crucial for effective learning. 
Our approach to sample video and text pairs $(v, t)$ of different lengths while requiring \textit{temporal overlap} improves video-text relevance and encourages fine-grained association.
As such, a video (or text clip) can have a better chance to be aligned or supervised by nearby text and vice versa. By contrast, video clips without any temporally aligned text are never contributing as a positive video-text pair in our objective.


\subsection{Retrieval Augmented Training}
\label{sec:retri}
Our intention is to learn to model more fine-grained video-text similarity by using difficult examples in our contrastive pre-training objective \eqref{eq:nce}.
We construct \textit{negatives} in our training batch by using hard pairs  $\{z_t^{-}\}$, which are semantically to the pairs in the numerator, using retrieval based sampling.


Recall that contrastive loss (\eg in equation \eqref{eq:nce}) uses positive pairs in a batch $B$, and typically negative pairs are implicitly induced from other positive pairs in the same batch.

\noindent \textbf{Dense Video Cluster Retrieval.}
Our approach aims to find video clusters to construct a batch of training samples.
We formulate this as a dense retrieval process on the latent space of a video,  derived from the video/text embeddings of our transformer that is trained by the contrastive loss~\eqref{eq:nce}.

Our overall training process can be described as a two-stage method that alternatively performs \textit{retrieval} and \textit{training} in each epoch, and is summarized in Algorithm~\ref{alg:simulator}.

\begin{algorithm}[!t]
\LinesNumbered
\DontPrintSemicolon
\caption{Retrieval Augmented Training}
\label{alg:simulator}
\SetKwInOut{Input}{Input} 
\SetKwInOut{Output}{Output} 
\SetKwProg{Def}{def}{:}{}
\Input{$\mathcal{V}$ is video set; $M$ is model.}
\BlankLine

\ForEach{epoch}{
    infer global features for all videos $\mathcal{V}$ on $M$: each video $V\in\mathcal{V}$'s global feature is computed as $z_V=\frac{1}{2|B_V|}\sum_{(v, t) \in B_V} (z_v + z_t)$, where $B_V$ indicates all clip pairs of $V$;\;
    build dense index on all videos' $z_V$;\; 
    retrieve $|\mathcal{C}|$ video clusters, where each cluster $c\in\mathcal{C}$ is sampled as $c \sim k\text{NN}(z_V, 2k)$, $|c| = k$ from a random video $V$;\;
    sample overlapped video-text pairs from $c\in \mathcal{C}$ to train $M$.\;
}
\end{algorithm}

For each epoch, Line 2-4 corresponds to the retrieval stage and Line 5 corresponds to the training stage. Specifics are as folows. 

Line 2 computes the global features $z_V$ for each video by averaging the embeddings of all  of its video-text clips. An ablation (in appendix) shows that this is better than using the starting clip of a video to infer the representative video embedding.

Line 3 constructs the dense index\footnote{We use FAISS: \url{https://github.com/facebookresearch/faiss}.} for all videos to be used in our retrieval-based training. 

Line 4 first finds $|\mathcal{C}|$ (corresponds to the number of overall batches in the training set) random videos, where each video $V$ yields a video cluster $c$ as follows.
We sample $|c|$ videos from $k$ neighboring videos of $V$. Instead of searching $k$ nearest videos directly (see ablation in Table~\ref{tbl:abl_retri}), we sample $k$ videos from the $2k$ nearest videos.
This is because we want videos in a cluster to be mutually closer to each other (not all close to video $V$). 
In this way, all video/text clips sampled from one video can serve as negative examples for clips sampled from another video.


\section{Zero-shot Transfer to End Tasks}
\label{sec:zs}
We present methods for zero-shot transfer of \mbox{VideoCLIP} to a variety of end tasks (\textit{without} using any labels).
For each task, we specify requirements that highlight the aspect of pre-training.

\noindent\textbf{Text$\rightarrow$Video Retrieval.}
Text$\rightarrow$video retrieval tests the text-to-video similarity computed on the learned video-text representation. 
$\text{NCE}(z_t, z_v)$ in Equation \ref{eq} contributes to this task as it discriminates different video clips in the numerator and denominator for a given text clip.
It also tests the distribution of hard negative examples in the denominator given it reports multiple recall metrics.

\noindent\textbf{Multiple-choice VideoQA.}
In multiple-choice VideoQA~\cite{yu2018joint}, the model aligns each video with one out of several text candidate answers.
It tests video$\rightarrow$text similarities with a pre-trained model. We formulate this task as ranking candidate textual answers for a given video question query.
This corresponds to the $\text{NCE}(z_v, z_t)$ term in Equation~\ref{eq}, where the subtle differences in texts are discriminated against each other.

\noindent\textbf{Action Segmentation.}
\label{sec:seg}
Action segmentation assigns each token (or frame) of a video with one of the pre-defined labels to separate meaningful segments of videos from the rest tokens (or frames).
This is similar to sequence labeling (\eg~named entity recognition (NER)) in NLP.
Inspired by the setup of CLIP~\cite{radford2021learning}, the text encoder of VideoCLIP can serve as self-supervision for videos during pre-training and as a hyper network to provide hidden states of segment textual labels for a video token.
As such, the hidden state of each video token can have a distribution of similarity over segment labels.
This task tests video token to text similarities.

One challenge in action segmentation is that it contains an \underline{O}utside label that does not exist in transcription during pre-training.
This \underline{O}utside label is task-dependent because it means a token does not belong to any of the pre-defined labels.
This is similar to open set recognition \cite{scheirer2012toward} or out-of-domain intent detection \cite{lane2006out}, where the \textit{rejection} label is not presented during training but all new classes during inference (not shown in training) should be covered by the \textit{rejection} label.

Let $t \in L$ be one label in the set of all labels $L$ excluding the \underline{O}utside label. 
We apply the following conditions to each video token $u$ to curate the prediction with the \underline{O}utside label $\hat{y}_u$:
\begin{equation}
\begin{split}
\begin{cases}
\argmax_{t\in L} (h_u z{_t}^T)&\text{if}\ \max_{t\in L}(h_u z{_t}^T)>\gamma, \\
\text{\underline{O}utside}&\text{otherwise,}
\end{cases}
\end{split}
\end{equation}
where $\gamma$ is a threshold.
Note that in zero-shot transfer, there is no access to training or validation data to decide a threshold as a hyper-parameter.
Thus, we estimate $\gamma$ as the maximum of dot products of intra-labels:
$\gamma = \max(z_{t} z_{t'}^T)$, where $t\in L, t'\in L$ and $t\neq t'$.

\noindent\textbf{Action Step Localization.}
In this task, each video is associated with a ``task'' with multiple steps $S$, where each step $t\in S$ is described as a short text.
Action step localization is to assign each video token to one or multiple steps in the associated task.
This is similar to action segmentation except that the label set is not pre-defined and does not contain the \underline{O}utside label.
As such, we first obtain the hidden states for each video frame (or token) $h_u$ from transformer. Then we separately forward text labels into the text backbone to obtain the hidden states of step labels $z_S$.
The distribution of each video token over steps is predicted as $\text{Softmax}(h_u z{_S}^T)$.

\section{Experiments}
\label{sec:exp}

\subsection{VideoCLIP Pre-training}
For pre-training, we use HowTo100M \cite{miech2019howto100m} that contains instructional videos via searching keywords from wikihow\footnote{\url{www.wikihow.com}} in YouTube. 
We use 1.1M videos after filtering out videos which are not available or cannot be decoded.
We randomly sample 4K videos as the validation set and use the rest for pre-training.
On average, the duration of each video is \app6.5 minutes with \app110 clip-text pairs.
After removing repeated words from ASR, we end up with \app7.7 GB of text transcriptions, with 2.4 tokens per second on average.

\subsection{End Task Setups}
\paragraph{Text$\rightarrow$Video Retrieval.} We use Youcook2, MSR-VTT and DiDeMo to evaluate zero-shot transfer to text-video retrieval.
Youcook2 \cite{zhou2017towards} has 2K cooking videos with a total duration of 176 hours and 5.26 minutes on average per video.
It shows about 89 recipes in 14K video clips.
Each video clip is annotated with one sentence. 
We follow the splits of~\citet{miech2019howto100m} to make sure there is no overlap between pre-training and evaluation data. 
We have 3,305 test clip-text pairs from 430 videos for zero-shot evaluation.
MSR-VTT~\cite{xu2016msr} is a well-known dataset for text-video retrieval, question answering etc.
Following JSFusion~\cite{yu2018joint,miech2019howto100m}, we randomly sampled 1K clip-text pairs as test data for evaluation of zero-shot transfer.
DiDeMo \cite{anne2017localizing} has 10,000 videos annotated with 40,000 sentences on Flicker videos. We evaluate video-paragraph retrieval on 4021 available testing examples\footnote{\url{https://github.com/LisaAnne/LocalizingMoments/blob/master/utils/eval.py}}.

\paragraph{VideoQA.} We further use the QA test data \cite{yu2018joint} for MSR-VTT to evaluate multiple-choice VideoQA.
Recall that this task can be formulated as a video-text retrieval task except the candidate textual answers are associated with each video and only one answer is correct (most relevant).
On average, VideoQA for MSR-VTT has 5 candidate answers per video.

\paragraph{Action Segmentation.} We use COIN \cite{tang2019coin} to evaluate action segmentation.
It has 11,827 videos (476 hours) in total and the testing set has 2797 videos, where each video is labeled with 3.91 segments per video on average.
There are 778 segment labels and we feed these textual labels into the text backbone to obtain their latent space.
As a reminder of Section \ref{sec:seg}, we do not model the \underline{O}utside label explicitly and determine an \underline{O}utside label only when all other 778 labels reject a video token. 
Note that videos in COIN can last for several minutes, we apply a sliding window with a step size of 16 seconds and a window size of 32 seconds.
During inference, we average the logits for overlapped tokens from multiple windows.

\paragraph{Action Step Localization.} We use CrossTask \cite{zhukov2019cross} to evaluate action localization. 
It contains 83 different tasks and 4.7K videos. 
Each task has a set of steps in the form of text descriptions and each frame of video is annotated with one or multiple steps as a distribution.
We use the testing data split via the official code\footnote{ \url{https://github.com/DmZhukov/CrossTask}}, which contains 1690 annotated videos.
We leave details of fine-tuning data to appendix.

\subsection{Implementation Details}
\paragraph{Video Encoder.}
We use a S3D~\cite{xie2018rethinking} for video encoder $f_{\theta_\text{CNN}}$. It is pre-trained on HowTo100M~\cite{miech2020end} to extract video tokens of dimension 512.
We use 30fps and extract one video token per second. This can be pre-computed for efficiency. 

\paragraph{Transformers.}
For the video and text Transformers, $f_{\theta_\text{v}}$ and $f_{\theta_\text{t}}$, we initialize their weights with the pre-trained $\text{BERT}_{\text{BASE-uncased}}$~\cite{devlin-etal-2019-bert}. Using the same type of transformer further allows us to perform ablation study on sharing video and text backbones (see Table \ref{tbl:abl_retri}).
We only use the first 6 Transformer layers for the video input and all 12 layers for the text input.
Please note that the video/text encoders in VideoCLIP is generally applicable to other pre-trained Transformers.
We use a single layer MLP $f_{\theta_\text{MLP}}$ with GELU activation~\cite{hendrycks2016gaussian} to map the S3D outputs to the 768-dimensional inputs of the video Transformer.


We limit the maximum number of video tokens to be 32.
For video transformer, its input sequence is 34 with \texttt{[CLS]} and \texttt{[SEP]} tokens.
For text transformer, we have 61 text tokens plus \texttt{[CLS]} and \texttt{[SEP]} tokens (63 in total).
The number of text tokens roughly doubling in the number of video tokens because text comes at \app2.4 tokens per second (on average) in the HowTo100M data, while our video tokens are extracted at 1 token per second. 
A text clip has a random length between 8 and 61 tokens, whereas a video clip has 3 to 32 seconds.
We sample 16 video/text pairs from each video and use $k{=}$32 videos to form batches of size $|B|{=}$512.


\paragraph{Training Details.}
We pre-train our model on 8 NVIDIA Tesla V100 GPUs (each with 32 GB memory) for 25 epochs using fp16 precision for \app1 day.
We use Adam \cite{kingma2014adam} as optimizer with betas of (0.9, 0.98), an initial learning rate of 5e-5, 1000 steps of warm-up, and a polynomial decay learning rate schedule. 
Gradients are clipped at 2.0. The softmax temperature in objective \eqref{eq:nce} is set to $\tau=1.0$.

\begin{table}[!t]

\centering
\setlength\tabcolsep{0.1pt}
\scalebox{.83}{
    \begin{tabular}{l c c c c}
    \textit{Youcook2} dataset & R@1 $\uparrow$ & R@5 $\uparrow$ & R@10 $\uparrow$ \\
    \shline
    \textsc{Supervised}\\
    HGLMM\cite{klein2015associating} & 4.6 & 14.3 & 21.6\\
    Coot\cite{ging2020coot} & 16.7 & 40.2 & 52.3\\
    UniVL (FT-Joint)\cite{luo2020univilm} & 22.2 & 52.2 & 66.2\\
    \textbf{VideoCLIP} (Fine-tuned) & \textbf{32.2} & \textbf{62.6} & \textbf{75.0} \\ \shline
    \textsc{Zero-shot}\\
    Random & 0.0 & 0.2 & 0.3 \\
    HowTo100M\cite{miech2019howto100m} & 6.1 & 17.3 & 24.8 \\
    MIL-NCE\cite{miech2020end} & 15.1 & 38.0 & 51.2 \\
    \textbf{VideoCLIP} (Zero-shot) & \textbf{22.7} & \textbf{50.4} & \textbf{63.1}\\

    \\
\textit{MSR-VTT} dataset & R@1 $\uparrow$ & R@5 $\uparrow$ & R@10 $\uparrow$ \\
\shline
    \textsc{Supervised}\\
     UniVL (FT-Joint) \cite{luo2020univilm} & 20.6 & 49.1 & 62.9\\
     ClipBERT \cite{lei2021less} & 22.0 & 46.8 & 59.9\\
     MMT \cite{gabeur2020multi} & 25.8 & 57.2 & 69.3\\ 
     Support Set\cite{patrick2021supportset} & 30.1 & \textbf{58.5} & \textbf{69.3}\\
     \textbf{VideoCLIP} (Fine-tuned) & \textbf{30.9} & 55.4 & 66.8 \\
     \shline
     
         \textsc{Zero-shot}\\
    Random & 0.1 & 0.5 & 1.0 \\
    HowTo100M\cite{miech2019howto100m} & 7.5 & 21.2 & 29.6 \\
    MIL-NCE\cite{miech2020end} & 9.9 & \textbf{24.0} & \textbf{32.4} \\
    SupportSet\cite{patrick2021supportset} & 8.7 & 23.0 & 31.1 \\
    \textbf{VideoCLIP} (Zero-shot) & \textbf{10.4} & 22.2 & 30.0\\
    
    \end{tabular}
}
\caption{\textit{Text$\rightarrow$video retrieval} on Youcook2 and VTT. }
\label{tbl:youcook}
\end{table}

\subsection{Main Results}
We evaluate VideoCLIP on various end tasks and compare it with other zero-shot and supervised methods that use labels on the target datasets. 

\paragraph{Text-video Retrieval.} The results on Youcook2 and MSR-VTT are shown in Table~\ref{tbl:youcook}. The result on DiDeMo is shown in Table~\ref{tbl:didemo}.

On Youcook2 (Table~\ref{tbl:youcook}, top), VideoCLIP shows impressive performance gains and has much better accuracy than traditional supervised methods. The zero-shot transfer performance is even close to the performance level of supervised baselines with pre-training.
With fine-tuning, VideoCLIP reaches state-of-the-art on Youcook2.

On MSR-VTT (Table~\ref{tbl:youcook}, bottom), VideoCLIP shows solid improvements but with a larger zero-shot to supervised gap than on Youcook2. The major reason could be domain shift from HowTo100M to MSR-VTT. The captions in MSR-VTT are more descriptive (\eg, ``a basketball player is playing basketball'' and are less likely to appear in the transcriptions of HowTo100M). 
After fine-tuning, VideoCLIP reaches state-of-the-art R@1. Note that this is achieved \textit{without using any supervised data} such as ImageNet or \textit{large-scale external data} (\textit{i.e.}, 65 million Instagram data) used by the second best method,  Support Set~\cite{patrick2021supportset}.

\begin{table}[h]

\centering
\setlength\tabcolsep{0.1pt}
\scalebox{1.}{
    \begin{tabular}{l c c}
    \textit{DiDeMo} dataset & R@1 $\uparrow$ & R@5 \\
    \shline
    \textsc{Supervised}\\
    S2VT \cite{venugopalan2014translating} & 11.9 & 33.6 \\ 
    FSE \cite{zhang2018cross} & 13.9 & 44.5\\ 
    CE \cite{liu2019use} & 16.1 & 41.1\\ 
    ClipBERT \cite{lei2021less} & 20.4 & 48.0\\ 
    \shline
    \textsc{Zero-shot}\\
    \textbf{VideoCLIP} (Zero-shot) & \textbf{16.6} & \textbf{46.9}\\ 
    \end{tabular}
}
\caption{\textit{Text$\rightarrow$video retrieval} on DiDeMo.}
\label{tbl:didemo}
\end{table}
On DiDeMo (Table~\ref{tbl:didemo}), VideoCLIP has better performance than most supervised methods. Note that ClipBERT\cite{lei2021less} has image pre-training before video+text fine-tuning.

\begin{table}[h]
\centering
\scalebox{0.83}{
    \begin{tabular}{l c}
\textit{MSR-VTT} dataset & Accuracy $\uparrow$ \\
\shline
\textsc{Supervised}\\
LSTM-fusion \cite{yu2018joint} & 38.3 \\
C+LSTM+SA-FC7 \cite{torabi2016learning} & 60.2 \\
SNUVL \cite{yu2016video} & 65.4 \\
EITanque \cite{kaufman2017temporal} & 65.5 \\
CT-SAN \cite{yu2017end} & 66.4 \\
VSE-LSTM \cite{kiros2014unifying} & 67.3 \\
MLB \cite{kim2016hadamard} & 76.1 \\
JSFusion\cite{yu2018joint} & 83.4 \\
ActBERT\cite{zhu2020actbert} & 85.7\\
ClipBERT\cite{lei2021less} & 88.2\\
\textbf{VideoCLIP} (Fine-tuned) & \textbf{92.1}\\
\hline
\textsc{Zero-shot}\\
\textbf{VideoCLIP} (Zero-shot) & \textbf{73.9}\\

    \end{tabular}
}
\caption{
\textit{VideoQA} on MSR-VTT.}
\label{tbl:videoqa}
\end{table}

\begin{table}[!htbp]
\setlength\tabcolsep{0.1pt}
\centering
\scalebox{0.8}{
    \begin{tabular}{l c}
\textit{COIN} dataset & Frame Accuracy $\uparrow$ \\
\shline
\textsc{Supervised}\\
NN-Viterbi \cite{richard2018neuralnetwork} & 21.2\\
VGG \cite{simonyan2014very} & 25.8\\
TCFPN-ISBA \cite{ding2018weakly} & 34.3\\
CBT \cite{sun2019contrastive} & 53.9\\
ActBERT \cite{zhu2020actbert} & 57.0\\ 
MIL-NCE \cite{miech2020end} & 61.0\\
\textbf{VideoCLIP} (Fine-tuned) & \textbf{68.7} \\
\shline
\textsc{Zero-shot}\\
\textbf{VideoCLIP} (Zero-shot) & \textbf{58.9} \\
    \end{tabular}
}
\caption{\textit{Action segmentation} on COIN.}
\label{tbl:coin}
\end{table}


\paragraph{Video Question Answering.} 
In Table~\ref{tbl:videoqa}, zero-shot VideoCLIP outperforms most supervised methods but similarly suffers from domain shift from HowTo100M to MSR-VTT.
After fine-tuning, it reaches the best performance, indicating VideoCLIP also provides strong features for fine-tuning.

\paragraph{Action Segmentation.} We report the results of action segmentation on COIN in Table~\ref{tbl:coin}. Zero-shot transfer of VideoCLIP to COIN outperforms all supervised methods, \textit{without using any labels on this dataset}.
This indicates that VideoCLIP also learns good token-level video representations.
Fine-tuning VideoCLIP further yields a \app10\% accuracy gain, indicating potential room for improvement.

\paragraph{Action Step Localization.}
Lastly, we report VideoCLIP's performance on CrossTask in Table~\ref{tbl:crosstask}.
It shows a small gap to supervised methods when using zero-shot action step localization. 
Fine-tuning leads to a \app10\% gain, outperforming all prior work on this dataset.

\begin{table}[!t]
\centering
\setlength\tabcolsep{2.0pt}
\scalebox{0.85}{
    \begin{tabular}{l c}
\textit{CrossTask} dataset & Average Recall $\uparrow$\\
\shline
\textsc{Supervised}\\
Alayrac \cite{alayrac2016unsupervised} & 13.3\\
Zhukov \cite{zhukov2019cross} & 22.4\\
Supervised \cite{zhukov2019cross} & 31.6\\
ActBERT \cite{zhu2020actbert} & 41.4\\ 
UniVL \cite{luo2020univilm} & 42.0\\
\textbf{VideoCLIP} (Fine-tuned) & \textbf{47.3}\\
\shline
\textsc{Zero-shot}\\
HowTo100M \cite{miech2019howto100m} & 33.6\\
MIL-NCE \cite{miech2020end} & 40.5\\
\textbf{VideoCLIP} (Zero-shot) & \textbf{33.9} \\
    \end{tabular}
}
\caption{\textit{Action step localization}  on CrossTask.}
\label{tbl:crosstask}
\end{table}

\begin{table*}[!htbp]
\small
\centering
\scalebox{0.85}{
    \begin{tabular}{l l l}
Query Text & Text of Top-1 video from \textbf{VideoCLIP} (Zero-shot) & Text of Top-1 video from \textbf{VideoCLIP} (Fine-tuned)\\
\shline
pick the ends off the verdalago & 
\shortstack[l]{put chickpeas parsley chopped onion chili powder\\ ground cumin in food processor}
 & pick the ends off the verdalago\\
 \hline
add the fried pita to the salad and mix & toss the salad & add the dressing and bread pieces the the salad\\
\hline
\shortstack[l]{place chicken in hot oil\\ and fry until golden brown} & fry the chicken in oil & fry the chicken wings in deep oil\\
\hline
\shortstack[l]{fry dark meats together and\\ white meats together} & add the mutton to the pan & add the diced beef meat to it and roast it\\
\hline
rub salt and pepper onto the chicken & season them with salt and pepper & rub salt and pepper onto the chicken\\
\hline
    \end{tabular}
}
\caption{Qualitative error analysis of \textit{text$\rightarrow$video retrieval} on Youcook2.
}
\label{tbl:err_youcook}
\end{table*}

\subsection{Discussion on Work that Fine-tunes CLIP Model}
There are concurrent works \cite{luo2021clip4clip,portillo2021straightforward} about using image+text model \cite{radford2021learning} for video+text downstream tasks.
Note that \cite{luo2021clip4clip} and \cite{portillo2021straightforward} use image pre-training (no video pre-training) and transfer to videos, whereas our focus is about improving video pre-training using a novel pre-training objective. 
Besides this conceptual difference
\cite{luo2021clip4clip,portillo2021straightforward} are using a pre-trained image CLIP\cite{radford2021learning} model from OpenAI which is trained on huge, semi-curated web image+text pairs that provides exceptional zero-shot performance on many datasets (\eg  ImageNet); however, the CLIP pre-training data is sourced from web-search engines (which on their own use fully supervised neural networks trained on ImageNet and other datasets); therefore, is not fair to compare to our approach which only trains on HowTo100M instructional videos.

\subsection{Ablation Study}
In Table~\ref{tbl:abl_retri},
we perform an ablation study on zero-shot transfer for text$\rightarrow$video retrieval on Youcook2 to quantify the 
the contribution of overlapping clips and retrieval augmented pre-training.

In the first group, we study the effectiveness of the two proposed methods.
VideoCLIP without retrieval augmented training significantly drops performance by over \textbf{4}\%  in R@1 and additionally using \textit{exact alignment} positives,~\textit{i.e.}, the same start/end timestamp for a pair of video and text clips, has another \textbf{4}\% drop in R@1. Therefore, both techniques combined lead to a \app50\% relative improvement in recall.

Further, by \textit{using MIL-NCE clips and loss} we evaluate the potential benefit of using the training objective from MIL-NCE~\cite{miech2020end} (which uses multiple temporally adjacent clips as positives) in our architecture. This ablation isolates the pre-training objective from model and data.
We observe that the MIL-NCE loss can improve the direct alignment objective but performs significantly worse than our objective (16.1 \textit{vs.}~22.7 R@1). 


%

\begin{table}[t!]
\centering
\setlength\tabcolsep{2.0pt}
\scalebox{0.84}{
    \begin{tabular}{l c c c}
    \textit{Youcook2} dataset & R@1 $\uparrow$ & R@5 $\uparrow$ & R@10 $\uparrow$ \\
    \shline
    \textbf{VideoCLIP} (Zero-shot) & \textbf{22.7} & \textbf{50.4} & \textbf{63.1}\\
    $-$ w/o retrieval & 18.5 & 42.8 & 54.6 \\
    $-$ w/o retrieval and w/o overlap & 12.4 & 30.2 & 40.7 \\ 
    $-$ using MIL-NCE clips and loss & 16.1 & 38.6 & 51.1 \\
    \shline
    $-$ shared video/text transformer & 21.9 & 48.1 & 60.6 \\
    $-$ retrieve $k$ & 22.5 & 49.3 & 61.4\\
    $-$ use first 32 sec for retrieval & 20.1 & 46.3 & 58.7\\
    $-$ use \texttt{[CLS]} & 22.1 & 47.1 & 59.6\\
    \end{tabular}
}

\caption{Ablation on \textit{text$\rightarrow$video retrieval} (Youcook2).
}
\label{tbl:abl_retri}
\end{table}

%
%

In the second group, we further study the design choices of modeling.
\textit{shared video/text transformer} indicates $ f_{\theta_\text{v}}$  is the same as $ f_{\theta_\text{t}}$, which only decreases performance slightly. This suggests that using a joint backbone for video and text is effective. 

\textit{retrieve $k$} indicates direct searching $k$ nearest neighbors instead of sampling $k$ videos from $2k$ nearest neighbors (used by VideoCLIP) in Line 4 of Algorithm 1.
Sampling from nearest neighbors yields video clusters of better quality.

\textit{use starting 32 sec for retrieval} indicates using the first 32 secs of a video as representation for video retrieval, which is an inferior representation of the whole video.

Unlike employing \texttt{Avgpool}, using \texttt{[CLS]} token only prevents VideoCLIP from exploiting token-level information and thus yields worse performance.

\subsection{Qualitative Analysis}

We examine errors for text-video retrieval of Youcook2 in both zero-shot transfer and fine-tuning setting in Table~\ref{tbl:err_youcook}.
We observe that in zero-shot transfer, VideoCLIP has no prior knowledge about a particular task/dataset on how long a text and video clip should be paired together for the text-retrieval task.
Fine-tuning allows to correct this type of error.
Further, we observe that VideoCLIP tends to mix objects of similar color/shape together.
We leave incorporating such type of knowledge into pre-training to future work.

\vspace{5pt}
\section{Conclusion}
We have presented VideoCLIP, an approach to pre-train a video-text model for zero-shot transfer to end tasks that require fine-grained association between video and language. 
VideoCLIP uses an objective that contrasts temporally overlapping positives with hard negatives stemming from nearest neighbor retrieval. 
In evaluation this approach outperforms prior work on a variety of tasks, without any supervision on downstream datasets, and in some cases VideoCLIP is competitive or better than prior work that uses full supervision; nevertheless, we still observe gains for fine-tuning our model. We hope that our code and model will foster future research in multi-modal video understanding. 

\vspace{5pt}
\section*{Code}

Code and models are made available at \url{https://github.com/pytorch/fairseq/tree/main/examples/MMPT}.

\vspace{5pt}
\section*{Acknowledgments}
We thank Licheng Yu for in-depth discussion and feedback, as well as Huaishao Luo and Mandela Patrick for supporting baseline implementation.




\bibliographystyle{acl_natbib}
\bibliography{emnlp2021}

\newpage
\appendix

\section{Supplementary Material for VideoCLIP}
This supplementary material is organized as follows.
First we provide additional experimental setups for each end task. 
Then we specify the hyper-parameters in our model and detail how we train VideoCLIP.
Lastly, we provide extra ablations and analysis of various VideoCLIP configurations.

\subsection{End Task Setup Details}
\noindent \textbf{Text-Video Retrieval.} We use Youcook2 and MSR-VTT to evaluate text-video retrieval. We directly use our video and text Transformers to encode the videos and the text queries and measure the text-to-video similarities for retrieval.

Youcook2 \cite{zhou2017towards} is a collection of 2K cooking videos with a total duration of 176 hours and 5.26 minutes on average per video.
It contains 89 recipes in 14K video clips where each clip is annotated with one descriptive sentence. 
We follow the splits defined in~\citet{miech2019howto100m} and make sure there is no overlap between pre-training and evaluation data. 
After filtering out unavailable ones, we obtain 9,473 training clip-text pairs from 1222 videos and 3,305 test clip-text pairs from 430 videos.

MSR-VTT~\cite{xu2016msr} is a widely-compared benchmark dataset for text-video retrieval and video question answering.
It contains open-domain videos where each video clips is around 10 seconds. Each training clip has 20 captioning sentences labeled by a human. 
In total, there are 200K clip-text pairs from 10K videos.
Following JSFusion~\cite{yu2018joint,miech2019howto100m}, we sampled 1K clip-text pairs as the test data and the rest is used for training.

\noindent \textbf{Multiple-choice VideoQA.} We use the testing split and data in~\cite{yu2018joint} on MSR-VTT to evaluate multiple-choice VideoQA.
On average, VideoQA for MSR-VTT has 5 candidate answers per video.
Recall that this task can be formulated as a video-text retrieval task except the candidate textual answers are associated with each video and only one answer is correct (most relevant).
In practice, we find the answer with the maximum similarity in-between a video and all candidate answers.

\noindent \textbf{Action Segmentation.} 
We use COIN \cite{tang2019coin} to evaluate action segmentation.
COIN contains 11,827 videos (476 hours) in total and the testing set has 2797 videos, where each video is labeled with 3.91 segments per video on average.
There are 778 segment labels and we feed these textual labels into the text backbone to obtain their latent space.
We do not model the \underline{O}utside label explicitly and determine an \underline{O}utside label only when all other 778 labels reject a video token. 
Note that videos in COIN can last for several minutes, we apply a sliding window with a step size of 16 seconds and a window size of 32 seconds.
During inference, we average the logits for overlapped tokens from multiple windows.
For follow the original split of COIN for training and evaluation.

\noindent \textbf{Action Step Localization.} CrossTask \cite{zhukov2019cross} is used to evaluate action localization. 
There are 83 different tasks and 4.7K videos where each task has a set of steps in the form of text descriptions and each frame of video is annotated with one or multiple steps as a distribution.
We use the testing data split and the official codebase (\url{https://github.com/DmZhukov/CrossTask}) that contains 1.7K videos.
We use 540 annotated videos for supervised training.
Recall that action step localization testing the video's token-level features and we use the representations $\bm{h}_v$ of the last layer of BERT before average pooling.
We compute the distribution of similarity for each token over the latent space of textual labels of steps.

\end{document}